\documentclass{article}

\usepackage{arxiv}

\usepackage[utf8]{inputenc} % allow utf-8 input
\usepackage[T1]{fontenc}    % use 8-bit T1 fonts
\usepackage{hyperref}       % hyperlinks
\usepackage{url}            % simple URL typesetting
\usepackage{booktabs}       % professional-quality tables
\usepackage{amsfonts}       % blackboard math symbols
\usepackage{nicefrac}       % compact symbols for 1/2, etc.
\usepackage{microtype}      % microtypography
\usepackage{lipsum}		% Can be removed after putting your text content
\usepackage{graphicx}
\usepackage{natbib}
\usepackage{doi}

\title{A Nature-Inspired Colony of Artificial Intelligence System with Fast, Detailed, and Organized Learner Agents for Enhancing Diversity and Quality}

\author{ \href{https://orcid.org/0000-0003-3235-9870}{\includegraphics[scale=0.06]{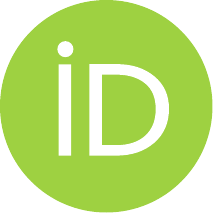}\hspace{1mm}Shan Suthaharan}\thanks{This is the authors’ preprint version of a paper accepted for publication in the Proc. of the AAAI Spring Symp. Series, 2025} \\
	Department of Computer Science\\
	UNC Greensboro\\
	Greensboro, NC 27402 \\
	\texttt{s\_suthah@uncg.edu} \\
	 \\
}

\renewcommand{\shorttitle}{A Nature-Inspired Colony of Artificial Intelligence}

\hypersetup{
pdftitle={A template for the arxiv style},
pdfsubject={q-bio.NC, q-bio.QM},
pdfauthor={David S.~Hippocampus, Elias D.~Striatum},
pdfkeywords={First keyword, Second keyword, More},
}

\begin{document}
\maketitle

\begin{abstract}
The concepts of convolutional neural networks (CNNs) and multi-agent systems are two important areas of research in artificial intelligence (AI). In this paper, we present an approach that builds a CNN-based colony of AI agents to serve as a single system and perform multiple tasks (e.g., predictions or classifications) in an environment. The proposed system impersonates the natural environment of a biological system, like an ant colony or a human colony. The proposed colony of AI that is defined as a role-based system uniquely contributes to accomplish tasks in an environment by incorporating AI agents that are fast learners, detailed learners, and organized learners. These learners can enhance their localized learning and their collective decisions as a single system of colony of AI agents. This approach also enhances the diversity and quality of the colony of AI with the help of Genetic Algorithms and their crossover and mutation mechanisms. The evolution of fast, detailed, and organized learners in the colony of AI is achieved by introducing a unique one-to-one mapping between these learners and the pretrained VGG16, VGG19, and ResNet50 models, respectively. This role-based approach creates two parent-AI agents using the AI models through the processes, called the intra- and inter-marriage of AI, so that they can share their learned knowledge (weights and biases) based on a probabilistic rule and produce diversified child-AI agents to perform new tasks. This process will form a colony of AI that consists of families of multi-model and mixture-model AI agents to improve diversity and quality. Simulations show that the colony of AI, built using the VGG16, VGG19, and ResNet50 models, can provide a single system that generates child-AI agents of excellent predictive performance, ranging between 82\% and 95\% of F1-scores, to make diversified collective and quality decisions on a task.
\end{abstract}

% keywords can be removed

\section{Introduction}

In artificial intelligence (AI) discipline, one of the emerging topics is the multi-agent AI systems that integrate the concepts of multi-agent adaptive learning systems \cite{boutilier1996planning, newton2021scalability}. The agents in a multi-agent AI system are generally unstructured and formed by complex functional and logical characteristics that are difficult to learn, understand, and interpret \cite{luzolo2024combining, foerster2018counterfactual}. The AI agents are goal-driven members of a multi-agent system; hence, their inherent communication structures are explicitly defined to form a globally inspired interactions and making decisions (e.g., predictions or classifications). Hence, a multi-agent AI system usually fails to efficiently integrate localized information to make collective decisions. In other words, they do not fully utilize the benefits of the concepts of a decentralized framework \cite{ferber1999multi}. Therefore, we need a multi-agent AI system that integrates families of AI agents that operate on a localized set of simple rules for collective decision making. Most recently, a concept of colony of AI has been proposed in AI research literature \cite{suthaharan2024colony}. This framework mimics biological colonies, like ants, bees, and humans \cite{dorigo2005ant}. Apparently, the colony of AI operates on a set of simple rules and generates families of AI agents to form a colony and make collective and quality decisions. 

\begin{figure*}[t!]
	\begin{centering}
	\includegraphics[scale=0.57]{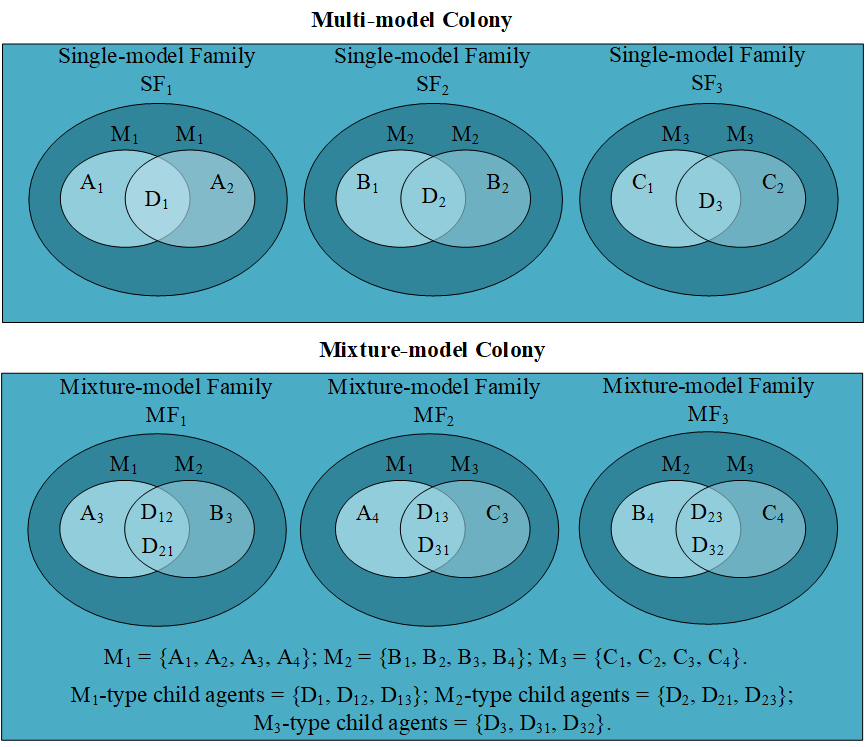}
	\caption{Overview of the concept of colony of AI.}
	\label{fig:colony}
	\end{centering}
\end{figure*}

The idea of this approach is to allow two AI agents, called the parent-AI agents, to share their knowledge (learned parameters) and produce child-AI agents. This process is called the marriage of AI agents, and it is performed using Genetic Algorithms (GAs) and their crossover and mutation mechanisms for the evolution of a colony of AI \cite{holland1992adaptation}. In essence, the marriage of AI refers to the process of knowledge sharing between parent-AI agents to produce a child-AI agent with a new knowledge. Hence it forms a family of AI agents and makes decentralized decisions in the colony of AI. One of the drawbacks of this foundational framework is that it is only a single-model framework of convolutional neural networks (CNNs). As such it restricts the generalization and adaptivity of an AI colony. As a single model it implements a pretrained VGG16 model that is built on the fundamental logic of a CNN model \cite{simonyan2014very, krizhevsky2017imagenet}. This single-model colony of AI is extremely useful in the era of multi-agent artificial intelligence, because of its simplicity, knowledge localization, and the ability to make collective decisions with simplified rules. However, to meet the requirements of a colony, it must be upgraded to a role-based framework, in addition to a rule-based framework. 

In this paper, we propose an approach to establish a colony of AI that consists of families of multi-model and mixture-model AI agents with an enhanced diversity and quality in the AI colony. This approach will promote the collective decision strategy and the adaptive intelligence approach in the colony \cite{stone2000multiagent}. Our proposed technique will include the families of fast learner AI agents, detailed learner AI agents, and organized learner AI agents to the colony and establish a multi-model colony of AI. In the multi-model colony of AI, a concept of intra-marriage of AI agents is defined to allow the sharing of knowledge between two agents of the same learner type, produce child-AI agents of the same learner type, and form families of AI agents. This approach enhances sample (localized) diversity \cite{hazra2022applications} and quality within each type of learners. The approach also includes mixture-models to establish families of mixture-model AI agents in the AI colony. A concept of intermarriage of AI agents is defined to allow the sharing of knowledge between two agents of different learner types, produce child-AI agents of multiple learner types and enhance diversity and quality in the proposed colony of AI framework. Thus, as a single system, the proposed colony of AI can control overfitting problems.  

\begin{figure*}[t!]
	\begin{centering}
	\includegraphics[scale=0.60]{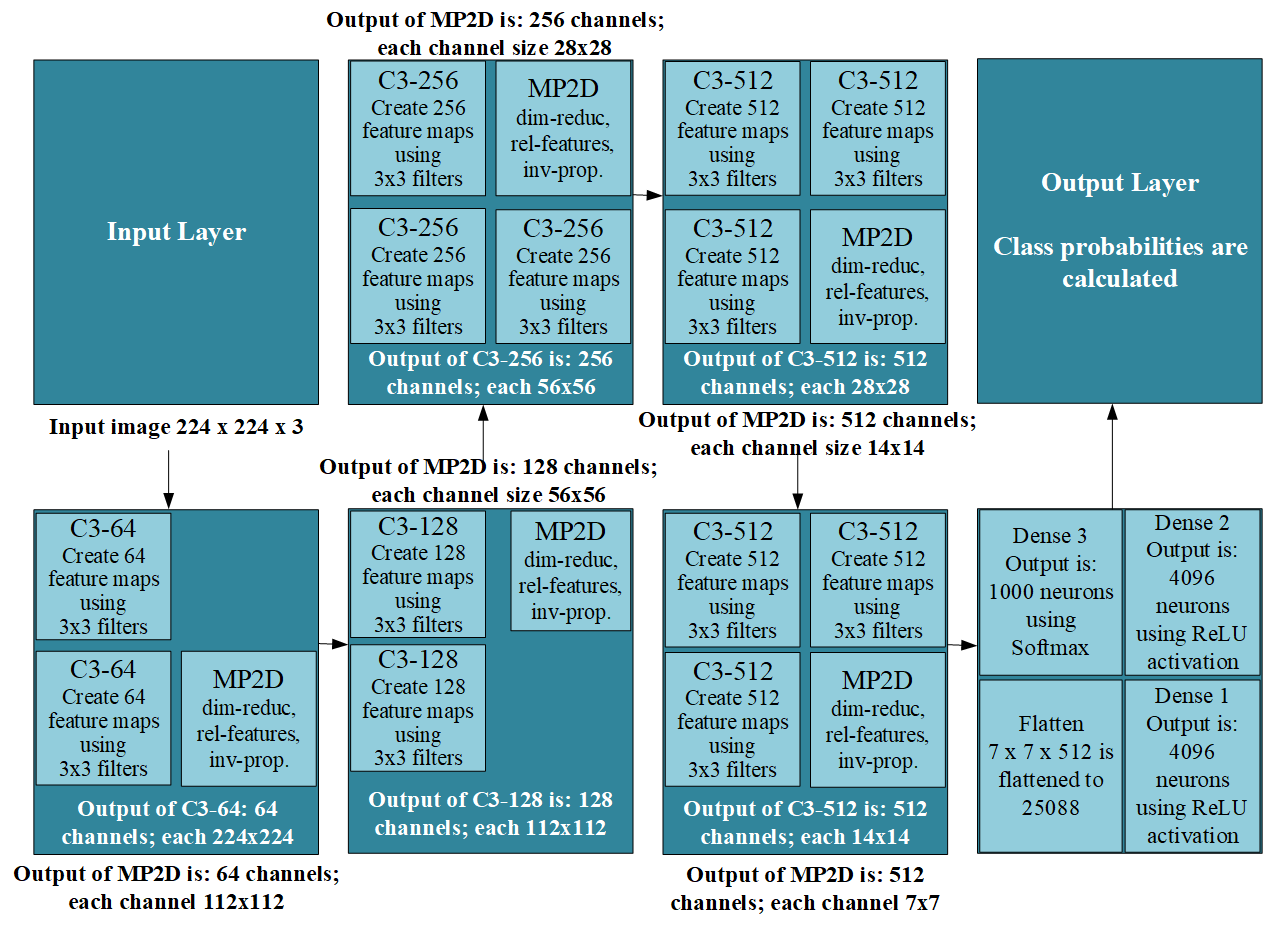}
	\caption{Fast learner: VGG16 architecture and the understanding of knowledge sharing by AI agents. Note: dim-reduc represents dimensionality reduction, rel-feature represents relevant features, and inv-prop represents invariant properties.}
	\label{fig:vgg16}
	\end{centering}
\end{figure*}

\section{Methods}

A colony of AI may be defined as follows: Suppose an AI agent $A_1$ is generated from a pretrained model $M_1$ (we may compare it with genetics) where the AI agent $A_1$ can perform a task $T_1$ in an environment $E$ as an agent of the model $M_1$. The environment $E$ generally can have distinct tasks; hence, the agent $A_1$ may be trained on a different task $T_2$ within the environment $E$. If the agent $A_1$ is retrained on a different task, then it becomes incapable of performing the previous task $T_1$ that was originally trained on. Therefore, the training of that nature may not be suitable for an environment that consists of many tasks and AI agents. To meet such conditions, we need a colony of AI that understands the local environment to collectively make decisions as a single AI system \cite{stone2010ad}. A scenario is presented in Figure \ref{fig:colony} that illustrates the proposed concepts of colony of AI. For example, as illustrated in this figure, let’s assume that we have two AI agents $A_1$ and $A_2$ of a pretrained model $M_1$, where the agents perform the same task $T_1$ by inheriting the behavior of the model $M_1$. Then the agents $A_1$ and $A_2$ go through the process of “marriage of AI” that adapts the operations of crossover and mutation of Genetic Algorithms. 

In this process, the model weights and bias parameters are shared, based on a probabilistic rule-based mechanism to produce a child-AI agent $D_1$ to form a single family $F_1$, as shown in Figure \ref{fig:colony}. Now the child-AI agent $D_1$ can be trained on a different task $T$, say $T \in \{T_1, T_2, \dots, T_p\}$, to expand the family to perform multiple tasks. Hence, the single-model approach delivers a mechanism to evolve the colony of AI as a single-model multi-agent families that mimics biological systems. However, the environment may experience issues related to the optimization of diversity and quality \cite{jaramillo2024understanding} of the colony of AI because of the heterogeneity, scalability, and the unstructured nature of the tasks that are generated by the large number of data sources in an environment. To address such complexity issues, we propose to establish a role-based approach, in addition to the rule-based approach. This role-based approach introduces fast learners, detailed learners, and organized learners among AI agents to build a colony of AI that satisfies the diversity and quality requirements while maintaining a path to optimality. Hence, the proposed approach combines three types of models: single-model colony of AI, multi-model colony of AI and mixture-model colony of AI that will make the colony more efficient and effective.   

\begin{figure}[t!]
	\begin{centering}
	\includegraphics[scale=0.62]{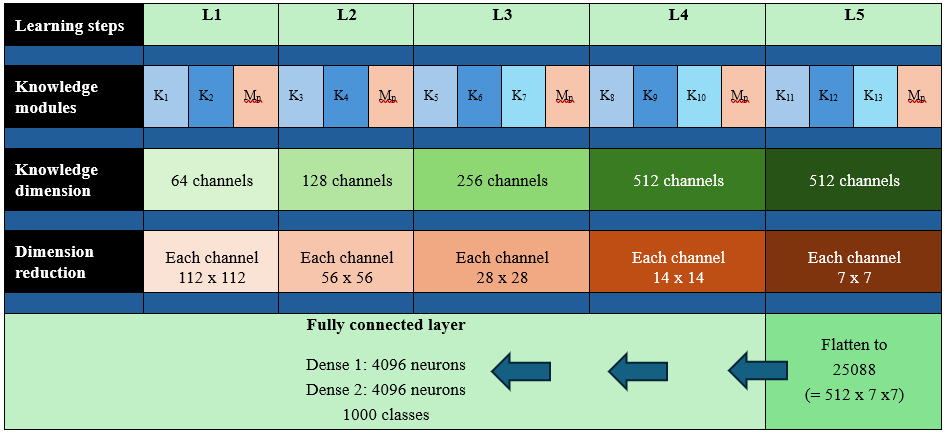}
	\caption{Pre-trained VGG16 internal architecture with knowledge mapping.}
	\label{fig:vgg16know}
	\end{centering}
\end{figure}

\subsection{Multi-model colony of AI}

It is defined as a colony that consists of multiple single-model families as shown in Figure \ref{fig:colony}: single-model family $SF_1$, single-model family $SF_2$, and single-model family $SF_3$. For example, single-model family $SF_1$, allows the intra-marriage between the agents of the same model type $M_1$, but in our proposed colony of AI framework, it represents a fast learner family of AI agents. Similarly, the single-model family $SF_2$ represents a detailed learner family of AI agents, and the single-model family $SF_3$ represents an organized learner family of AI agents. The goal of adding these three types of families is to increase diversity and quality of the colony of AI as a single system. In the proposed approach, we respectively generate a one-to-one mapping between VGG16, VGG19 \cite{simonyan2014very}, and ResNet50 \cite{he2016deep} and the fast, detailed, and organized learners. A family in the colony of AI is represented by 5-tuples $(p, q, r, s, t)$, where the parameters $p$ and $q$ represent parent-AI agents, and $r$, $s$, and $t$ represent child-agent, data-size, and training-duration, respectively.

\subsubsection{Fast learners}
The fast learners pay less attention to details while learning from data. Hence, they are good with data that are not that complex. They learn fast from simple data because the complex data may have many hidden details that can slow down their learning rate. Their goal is to complete the tasks as fast as they can with an effort to achieve good performance as much as possible. Therefore, in our approach, we map an AI agent derived from a VGG16 model to a fast learner AI agent (because of the simpler structure of VGG16) and represent this family by $(16, 16, 16, s, t)$, where $s=10,000$ data points (or images) and $t=3$ epochs. The architecture of a VGG16 model is presented in Figure \ref{fig:vgg16} that shows it has 13 convolutional layers, 5 max pooling layers, and 3 fully connected layers. In our proposed framework, we represent this architecture by a set of triplets $\{(1,2,64)$, $(2,2,128)$, $(3,3,256)$, $(4,3,512) $,$ ~(5,3,512)\}$. To highlight the functionality of this triplets representation that enables knowledge accessibility and knowledge sharing, the fast learner architecture is mapped to a learning sequence: learning steps, knowledge modules, knowledge dimension, and knowledge dimensionality reduction. Thus, the VGG16 model architecture is parameterized by the triplets representation $(a,b,c)$, where $a$ represents a convolutional learning step, $b$ represents a knowledge module, and $c$ represents the dimensionality of the knowledge. This representation is presented in Figure \ref{fig:vgg16know}.  

\begin{figure}[t!]
	\begin{centering}
	\includegraphics[scale=0.62]{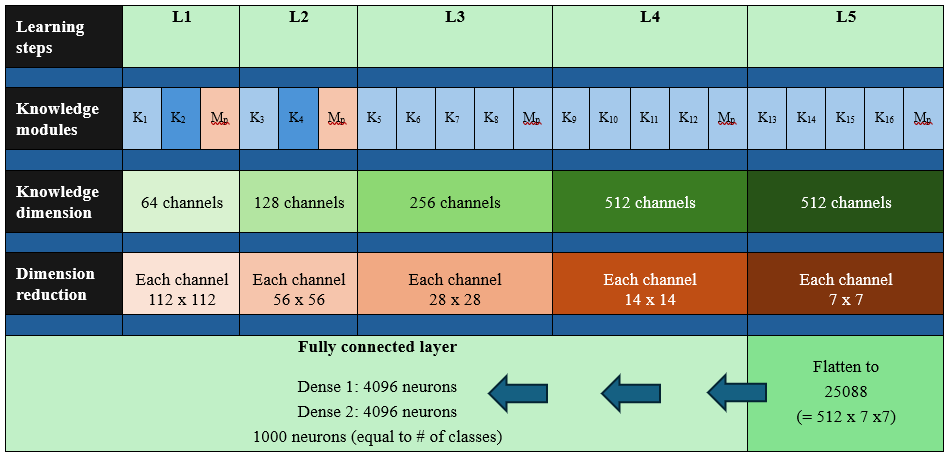}
	\caption{Pre-trained VGG19 internal architecture with knowledge mapping.}
	\label{fig:vgg19know}
	\end{centering}
\end{figure}

\subsubsection{Detailed learners}
They need more data and more time to learn. In other words, they are detail-oriented models; hence, they perform well with complex data than simple data. Thus, they require larger dataset than the one required by the fast learners to perform efficiently. Since VGG19 consists of 3 more convolutional layers than VGG16, we map an AI agent generated from VGG19 to a detailed learner and represent this detailed learner family by $(19, 19, 19, s, t)$, where $s \geq 10,000$ and $t=3$. The VGG19 architecture may also be described using a set of triplets $\{(1,2,64)$, $(2,2,128)$, $(3,4,256)$, $(4,4,512) $,$ ~(5,4,512)\}$, by following the same logic used to describe the fast learner architecture in Figure \ref{fig:vgg16}. Hence, the same parametrized triplets representation $(a,b,c)$ holds. In the set of triplets, for example, $(3,4,256)$ describes the third convolutional learning step, its fourth knowledge module, and 256 knowledge dimensionality. As such, the detailed learner architecture is also similarly mapped to the learning sequence: learning steps, knowledge modules, knowledge dimension, and knowledge dimensionality reduction. This representation is presented in Figure \ref{fig:vgg19know}.

\subsubsection{Organized learners}
\begin{figure}[t!]
	\begin{centering}
	\includegraphics[scale=0.62]{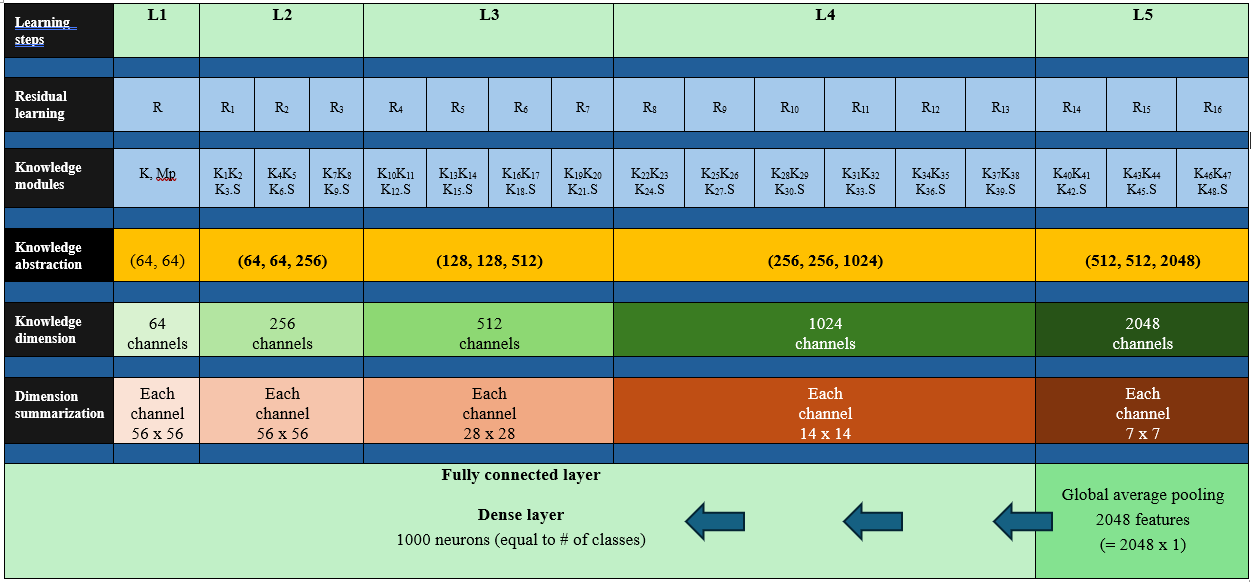}
	\caption{Pre-trained ResNet50 internal architecture with knowledge mapping.}
	\label{fig:resNet50know}
	\end{centering}
\end{figure}

The organized learners may not need more data to learn; however, they require more time to learn. In other words, they perform well with both complex and simple data through an organized structure that requires longer training time than other two types of learners. Since ResNet50 consists of more convolutional layers than VGG16 and VGG19 models and focuses on gradient, we can map an AI agent of ResNet50 to an organized learner and represent the family by $(50, 50, 50, s, t)$, where $n=10,000$ and $t=7$. The ResNet50 architecture may be described by combining a triplet and a set of hierarchical triplets as follows: $\{(1,2,64)$, $(2,1\dots3,(64,64,256))$, $(3,1\dots4,(128,128,512))$, $(4,1\dots6,(256,256,1024))$, $(5,1\dots3,(512,512,2048))\}$. Hence, the ResNet50 model architecture is parameterized using the hierarchical triplets representation $(a,b,(c,d,e))$, where $a$ represents a convolutional learning step, $b$ represents a residual learning step within a convolutional learning step, and $(c,d,e)$ represents the dimensionalities of knowledge module as a triplet, $c$, $d$, and $e$. Therefore, for example, to access the second knowledge module of the second residual learning step at the fourth convolutional learning step, we represent it by $(4,2,(-,256,-))$. Therefore, 
in contrast to VGG16 and VGG19, the organized learner architecture is mapped to the learning sequence: learning steps, residual learning, knowledge modules, knowledge abstraction, knowledge dimension, and knowledge dimensionality reduction (or summarization). This representation is presented in Figure \ref{fig:resNet50know}. As described earlier, this representation is also facilitate the knowledge accessibility and sharing to select the correct weights based on the dimensionality of knowledge.

\subsubsection{Intra-marriage of AI}
The intra-marriage of AI refers to the process of knowledge sharing between the parent-AI agents of the same learner type to produce a child-AI agent of the same learner types with a new knowledge. The concept of marriage of AI agents (intra or inter) first modifies the fully connected layer such that it supports the learning logic (fast, detailed, and organized) and prepares parent-AI agents to produce a child-AI agent to perform a new task. This process helps establish a new memory for a child agent. In addition, the other layers, as defined in previous sections, of the parent models carry their intermediate learning steps and knowledge modules with their weights and biases. Hence, these learning experiences will be transferred to the child-AI agent by sharing these parameters and forming a family of a learner type (fast, detailed, or organized).  

\subsection{Mixture-model colony of AI}

The purpose of mixture-model colony of AI is to enhance diversity and quality of the colony of AI by proposing the concept of intermarriage between the agents of distinct models. In a mixture-model colony of AI (Figure \ref{fig:colony}), the differences in the model architecture between the parent models create challenges in finding a mechanism for knowledge sharing. This is where, our triplet and hierarchical triplets representation contributes. An advantage of mixture model is that it automatically permits to have two types of child-AI agents based on the distinct types of the parent-AI agents. This feature helps increase the diversity and quality of the AI colony. 

\subsubsection{Fast learner and Detailed learner}
The marriage between a fast learner AI agent and a detailed learner AI agent brings a problem that is introduced by the requirement of a larger datasets by the detailed learners. The sharing of learned weights and biases should also be carefully addressed to overcome the problems induced by the misalignment of layers of these two distinct models and the learning strategy (vanishing gradient) \cite{hochreiter1998vanishing}. In this scenario, the parent-AI agents can have two child-AI agents and contribute to the increased diversity and quality in the AI colony. A fast learner parent-AI agent and a detailed parent-AI agent share their knowledge via the concept of intermarriage of AI agents, and produce a fast learner child-AI agent and a detailed learner child-AI agent. Therefore, with this concept and the proposed triplet knowledge representation, the parent-AI agents can share their knowledge modules, for example, $(3,3,256)$ of VGG16, and $(3,4,256)$ of VGG19 and $(5,3,512)$ of VGG16 and $(4,4,512)$ of VGG19 with the rule of 50\% (probabilistic rule) knowledge sharing (GA crossover) to produce child-AI agents (GA mutation).

\subsubsection{Fast learner and Organized learner}
The marriage between a fast learner AI agent and an organized learner AI agent brings different challenges that are imparted from the requirement of longer training time by the organized learners. Once again the sharing of learned knowledge modules should be carefully addressed to overcome the problems caused by the misalignment of layers of these two distinct models. The parents can also have two child-AI agents and increase diversity and quality, since they can produce one fast learner and one organized learner child-AI agents. 

A fast learner parent-AI agent and an organized learner parent-AI agent also share their knowledge via the intermarriage concept, and produce a fast learner child-AI agent and an organized child-AI agent. Hence, with this concept and the triplet and hierarchical triplet knowledge representations, the parent-AI agents can share their knowledge modules, for example, $(4, 2, (-,256,-))$ of ResNet50 and $(3,3,256)$ of VGG16, and $(5,2,(-,512,-))$ of ResNet50 and $(5,3,512)$ of VGG16 with the previously used rule of 50\% knowledge sharing to produce child-AI agents.

\subsubsection{Detailed learner and Organized learner}

In this case a detailed learner and organized learner parent-AI agents also share their knowledge through the concept of intermarriage of AI agents to produce a detailed child-AI agent and an organized child-AI agent. Therefore, with this concept and the proposed triplet and hierarchical knowledge representations, the parent-AI agents can share their knowledge modules $(4, 2, (-,256,-))$ of ResNet50 and $(3,4,256)$ of VGG19, and $(5,2,(-,512,-))$ of ResNet50 and $(4,4,512)$ of VGG19 with the 50\% sharing rule (crossover by GA) and produce child-AI agents by using mutation operation of GA.

\begin{table*}[ht]
%\caption{F1 scores of the models to predict each class}
\centering
\caption{F1 scores of the models to predict each class from the first set of simulations}
\vspace{0.2cm}
\begin{tabular}{|c|ccccccccccc|}
\hline
Family & 0 & 1 & 2 & 3 & 4 & 5 & 6 & 7 & 8 & 9 & Avg.\\[3pt]
\hline
\hline

F(16, 16, 16, 10k, 3) & 0.95 & 0.98 & 0.92 & 0.98 & 0.95 & 0.95 & 0.92 & 0.90 & 0.88 & 0.86 & 0.93\\ 
F(19, 19, 19, 10k, 3) & 0.49 & 0.98 & 0.85 & 0.94 & 0.94 & 0.63 & 0.73 & 0.93 & 0.58 & 0.92 & 0.82\\ 
F(50, 50, 50, 10k, 7) & 0.97 & 0.99 & 0.98 & 0.98 & 0.98 & 0.93 & 0.97 & 0.95 & 0.85 & 0.93 & 0.95\\ 
F(16, 19, 16, 10k, 3) & 0.95 & 0.98 & 0.91 & 0.95 & 0.97 & 0.87 & 0.94 & 0.95 & 0.85 & 0.93 & 0.93\\ 
F(16, 16, 19, 10k, 3) & 0.96 & 0.95 & 0.85 & 0.77 & 0.93 & 0.70 & 0.86 & 0.90 & 0.77 & 0.93 & 0.87\\ 
F(16, 50, 16, 10k, 3) & 0.95 & 0.99 & 0.89 & 0.87 & 0.96 & 0.91 & 0.85 & 0.92 & 0.90 & 0.91 & 0.92\\ 
F(16, 50, 50, 10k, 7) & 0.88 & 0.86 & 0.84 & 0.90 & 0.62 & 0.83 & 0.87 & 0.89 & 0.91 & 0.68 & 0.82\\ 
F(19, 50, 19, 10k, 3) & 0.80 & 0.98 & 0.91 & 0.84 & 0.95 & 0.77 & 0.80 & 0.95 & 0.85 & 0.93 & 0.89\\ 
F(19, 50, 50, 10k, 7) & 0.96 & 0.98 & 0.86 & 0.78 & 0.93 & 0.86 & 0.96 & 0.91 & 0.94 & 0.89 & 0.90\\ 
\hline
\end{tabular}
\label{tab:diversity}
\end{table*}

\subsection{Diversity-Quality trade-off estimator}

The trade-off between diversity and quality is an important factor to consider for measuring the effectiveness of a colony of AI agents as a single system. Our proposition is to use the combined effect of pairwise disagreement score, system entropy score, accuracy variance score, Mean Kendall's Tau and standard deviation of Kendall's Tau on the grid of accuracy values of the models and the number of labels to predict. Let us assume that we have $p$ models $\{a_1, a_2, \dots, a_p\}$ and $k$ labels $\{r_1, r_2, \dots, r_k\}$. Then we have a grid of $p \times k$ accuracy (F1-scores) values for the predictive performance of $p$ models for each label $r_i \in \{r_1, r_2, \dots, r_k\}$ . For example, in our approach, we have 9 models and 10 hand-written digits (labels 0 to 9) to predict (e.g., Table \ref{tab:diversity}) the digits.

To make the decision of diversity is much stronger, we also use Kernel density estimation (KDE) \cite{wkeglarczyk2018}. We know the accuracy is evolved through many random processes, model parameters, model optimization processes, switching of learned parameters (weight and biases); hence, we can assume each accuracy values is a random variable. Therefore, our goal is to estimate their probability density function and generate their corresponding kernel density estimators. This process confirms the origin of the accuracy values; hence, it's use can provide a meaningful and generalized interpretation for the diversity and quality trade-off. 

\section{Results}

Simulations have been conducted to validate the proposed concept of colony of AI that consists of both multi-model families of AI agents and mixture-model families of AI agents. A subset of the well-known MNIST dataset \cite{lecun1998gradient} and the VGG16, VGG19, and ResNet50 models for this validation purpose. We have constrained the MNIST dataset to 10000 images (or data points) for training and 2000 images for testing and analyzed the predictive performance of the models in terms of predicting 10 hand-written digits in the MNIST dataset. 

\subsection{Results from multi-model colony of AI}
This section presents the results from the simulations that analyze the performance of the proposed multi-model colony of AI agents that consists of the families of fast learner, detailed learner, and organized learner agents associated with the VGG16, VGG19 and ResNet50 models, along with the constrained-MNIST dataset.

\subsubsection{Intra-marriage VGG16 AI agents.}
\begin{figure*}[t!]
	\begin{centering}
	\includegraphics[scale=0.61]{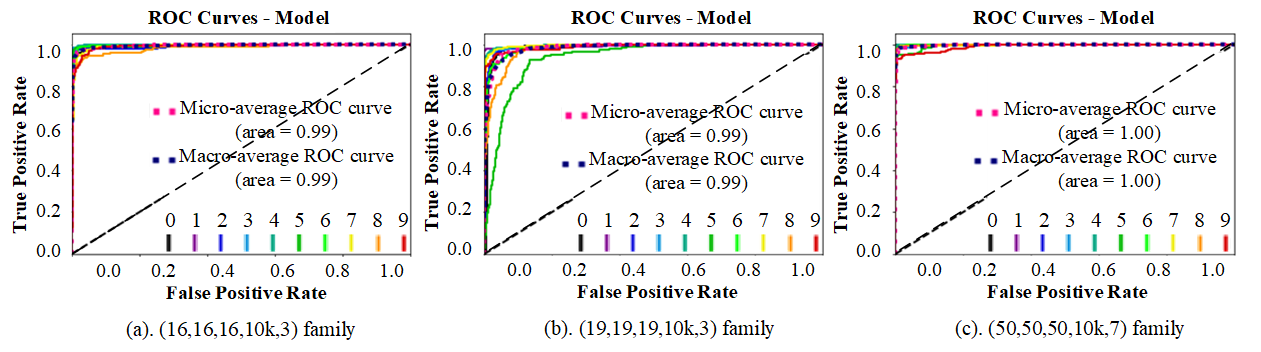}
	\caption{The results of ROC curves for the multi-model family of AI from the second set of simulations.}
	\label{fig:mult}
	\end{centering}
\end{figure*}

This simulation demonstrates that the child-AI agents produced by two VGG16 (fast learner) parent-AI agents display strong predictive performance on a constrained-MNIST dataset, even when more detailed of the data characteristics are not available. The F1-scores for this family are presented in the first row of Table \ref{tab:diversity} as performance measures \cite{suthaharan2016machine, powers2020evaluation}. These values show that the model (16,16,16) display very good performance in predicting the digits. In addition, the Receiver Operating Characteristic (ROC) curve presented in Figure \ref{fig:mult}(a) also supports that the model will give an excellent performance probabilistically, if additional data points are provided \cite{fawcett2006introduction, powers2020evaluation}.

\begin{figure*}[t!]
	\begin{centering}
	\includegraphics[scale=0.63]{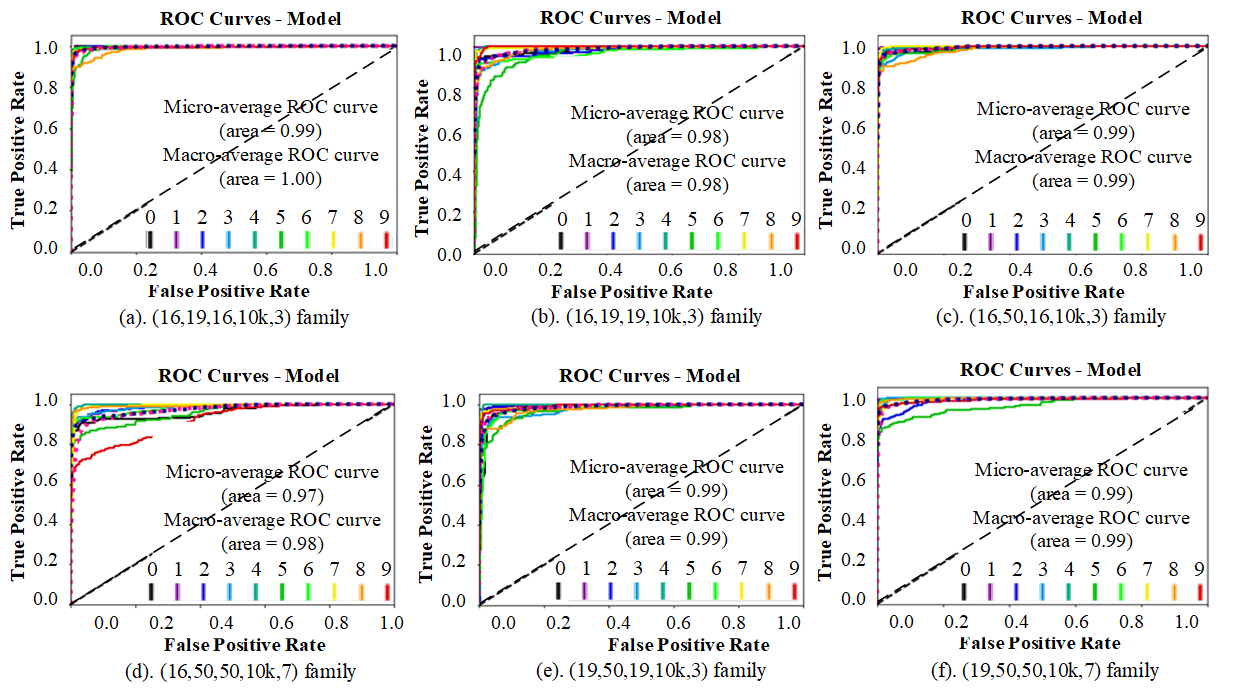}
	\caption{The results of ROC curves for the mixture-model family of AI from the third set of simulations.}
	\label{fig:mix}
	\end{centering}
\end{figure*}

\subsubsection{Intra-marriage VGG19 AI agents.}
We can observe from this simulation that the child-AI agents produced by two VGG19 (detailed learner) parent-AI agents may perform strong on some digits, but not for all. The values of the F1-scores for this family are presented in the second row of Table \ref{tab:diversity}. This is expected, as we mentioned before, because the extra layers of a VGG19 model (than VGG16 model) need additional data to extract detailed information for learning. We can also see from ROC curves presented in Figure \ref{fig:mult}(b) that the model can show excellent performance.

\subsubsection{Intra-marriage ResNet50 AI agents.}
We have also conducted a simulation to show that the child-AI agent produced by two ResNet50 (organized learner) parent-AI agents can give significantly higher performance than the other two models. The F1-scores of this family are presented in the third row of Table \ref{tab:diversity}. However, it requires more training time. In other words, we need to increase the number of epochs so that the ResNet50 models can focus more on the learning behavior (gradient). It means that ResNet50 requires additional time to address the vanishing gradient problem. The ROC results in Figure \ref{fig:mult}(c) show excellent performance.

\subsubsection{Collective performance.}
In summary, by considering the F1-scores in the first, second, and third rows of Table \ref{tab:diversity}, we can determine that the predictive performance of these models is diversified with good quality. The values show that the presence of model diversity in a colony of AI supports the task accomplishment. As an example, the low prediction accuracy of VGG19 on digit 0 does not matter to the overall performance of the colony since the collective decision suggests that an excellent prediction digit 0 can be achieved by VGG16 and ResNet50. Similar pattern can be seen in the prediction of digits 5 and 8. However, we observed separately that the performance of VGG19 can be improved to the level of VGG16 and ResNet50 with extra 2000 images.

\subsection{Results from mixture-model colony of AI}
This section presents the results of a set of simulations that analyze the performance of the colony of AI agents that consists of the mixture-model families of fast (VGG16) and detailed (VGG19) learners, fast (VGG16) and organized (ResNet50) learners, and detailed (VGG19) and organized (ResNet50) learners with the same constrained-MNIST dataset that was used in the previous simulations.

\subsubsection{Intermarriage, fast and detailed learners.}
The purpose of this simulation is to show that the child-AI agents produced by the fast and the detailed parent-AI agents through the concept of intermarriage display very good performance. These parent-AI agents can produce a fast learner child-AI agent (VGG16) and a detailed learner child-AI agent (VGG19). The F1-scores of these child-AI agents are presented in the fourth and fifth rows of Table \ref{tab:diversity}. By comparing these scores and the scores in rows one and two, we can see that the performance of VGG16 child is still very strong, while the performance of VGG19 has been improved because of the intermarriage between the fast and the detailed learners. In addition, the ROC curves in Figures \ref{fig:mix}(a) and (b) also support that the models can achieve excellent performance with additional data points.

\subsubsection{Intermarriage, fast and organized learners.}
This simulation is to show that the child-AI agents produced by the fast and organized parent-AI agents, through the concept of intermarriage, perform very good but with some deficiencies. In this scenario, the parent-AI agents produce a fast learner child-AI agent (VGG16) and an organized learner child-AI agent (ResNet50). The F1-scores of these child-AI agents are presented in the sixth and seventh rows of Table \ref{tab:diversity}. If we now compare these scores with the scores in rows one and three, we can say that the performance of the organized learners is slightly negatively influenced by the fast learner through intermarriage between the fast and organized learner agents. From the ROC curves presented in Figures \ref{fig:mix}(c) and (d), we can determine that excellent models' performance can be achieved with additional epochs.

\subsubsection{Intermarriage, detailed and organized learners.}
This simulation is to show that the child-AI agents produced by the detailed and organized parent-AI agents, through the concept of intermarriage, also perform very good. In this scenario, the parent-AI agents produce a detailed learner child-AI agent (VGG19) and an organized learner child-AI agent (ResNet50). The F1-scores of these child-AI agents are presented in the eighth and ninth rows of Table \ref{tab:diversity}. If we now compare these scores with the scores in rows two and three, we can say that the performance of the detailed and organized learners is balanced. It is clear from the ROC curves in Figures \ref{fig:mix}(e) and (f) that we can achieve excellent models' performance with additional data points and epochs.

\begin{figure}[t!]
	\begin{centering}
	\includegraphics[scale=0.76]{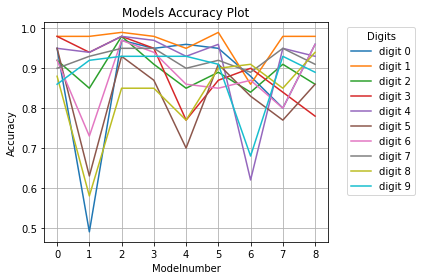}
	\caption{The diversity and quality plots for the models.}
	\label{fig:dq}
	\end{centering}
\end{figure}

\subsubsection{Comprehensive analysis.}
A set of results for comprehensive analysis is presented in Table \ref{tab:comprehensive}. It provides average F1 scores for training, validation, test performance, and the training time (epochs) of the 9 child-AI models, resulted from the intra- and inter-marriages between AI agents. We can see that the intra-marriage between two organized learners (ResNet50) that produces an organized child-AI agent (ResNet50) gives the best performance. However, overall the models' performance are very good (over 82\%), indicating diversity and quality. As discussed before, organized learners also need extra learning time to perform better. Note that we can achieve very high performance on all models with an increase in the number of data points and the training time.

\subsection{Diversity and quality}
Figure \ref{fig:dq} shows the visual representation of the data in Table \ref{tab:diversity}. It shows quality of the predictive behavior of the 9 inherited child-AI models using their prediction accuracy for the 10 digits (0 to 9) in 10 colors. Although we see diversity in the results of the models, it is essential to have numerical descriptors for the diversity. Hence we calculated the pairwise disagreement score (0.085), system entropy score (2.30), accuracy variance Score (0.007), mean Kendall's Tau (0.25), and standard deviation of Kendall's Tau: (0.27). The accuracy variance score of 0.007 indicates that the models display diversified performance in predicting digits \cite{hansen1990neural}; hence, it can contribute to the collective decision-making strategy. The mean Kendall's Tau value of 0.25 indicates that the models are not uniformly favoring the digits in predictive results \cite{kendall1938new}; hence, they show diversity behavior. These patterns can be seen in Figure \ref{fig:dq} which we can now interpret numerically. Similarly other measures also supporting diversity and quality of the models. We also used the KDE approach with exponential kernel and obtained the following results: pairwise disagreement score (0.084), system entropy score (2.28), accuracy variance Score (0.007), mean Kendall's Tau (0.58), and standard deviation of Kendall's Tau: (0.35). These results of a non-parametric approach also show similar diversity. 

\begin{table}[h]
%\caption{The performance results for different AI families}
%\vspace{0.1cm}
\centering
\caption{The performance results for different AI families}
\vspace{0.1cm}
\begin{tabular}{|ccccc|}
\hline
Models        & Train. & Valid. & Test & Duration \\
\hline
\hline
(16,16,16,10k,3)       & 0.94      & 0.95   & 0.93     & 443 sec.     \\
(19,19,19,10k,3)       & 0.79      & 0.83   & 0.82     & 592 sec.      \\
(50,50,50,10k,7)       & 0.99      & 0.96   & 0.95     & 713 sec.     \\
(16,19,16,10k,3)       & 0.95      & 0.95   & 0.93     & 442 sec.      \\
(16,19,19,10k,3)       & 0.83      & 0.88   & 0.87     & 591 sec.      \\
(16,50,16,10k,3)       & 0.91      & 0.94   & 0.92     & 440 sec.     \\
(16,50,50,10k,7)       & 0.97      & 0.84   & 0.82     & 724 sec.      \\
(19,50,19,10k,3)       & 0.82      & 0.90   & 0.89     & 613 sec.     \\
(19,50,50,10k,7)       & 0.96      & 0.92   & 0.90     & 732 sec.      \\
\hline
\end{tabular}
\label{tab:comprehensive}
\end{table}

The ultimate goal of measuring solution diversity is to collectively use the diversity scores to achieve optimality by treating the colony as a single system and reduce decision complexity in the environment (the problem domain). Also note that many diversity components contribute to the solution diversity. The first one is the role-based diversity that is achieved by integrating fast, detailed, and organized AI learners and assigning them to convolutional neural network models in this research. Similarly, the 5-tuples representation$(p, q, r, s, t)$ that is used to establish family history consists of (i) the role-based diversity that comes from the model types. $(p, q, r)$, of parents and child AI agents, (ii) the knowledge diversity $s$ that describes the origin of data (handwritten digits data, plant leaves, or medical data) and the amount of data that come from the environment, and (iii) the temporal diversity parameter $t$ that defines the training time and the suitability (lifetime) of the model over time. The proposed nature-inspired colony of AI also considers three more diversity parameters: learning diversity, intelligence diversity, and behavioral diversity. These diversities come from the triples and hierarchical representations, and the AI models' parameters -- the learning parameters, hyperparameters, learning rates, optimization functions, activation functions, and learned weights and biases. 

Hence, the proposed nature-inspired colony of AI brings many innovative advantages to the advancement of artificial intelligence that include:

\begin{itemize}
\item The mapping of the well-known AI models, like VGG16, VGG19, and ResNet50 to specific learning roles, like fast, detailed, and organized AI learners. As we have seen, it brings the role-based diversity to the colony of AI. The 5-tuples representation brings the combined role-based, knowledge-based, and temporal-based diversities. It also the inclusion of other models incrementally to enhance these diversities.

\item The triplets and hierarchical triplets representations provide technical tools that enable knowledge accessibility and knowledge sharing between AI agents. Through these structural, functional, and logical characteristics, they bring learning diversity, intelligence diversity, and behavioral diversity to the colony of AI.

\item The application of genetic algorithms with crossover and mutation mechanisms provides a powerful and flexible framework to generate a diversified colony with evolutionary intelligence. This process, in essence, enable the solution diversity to evolve and reduce decision complexity. 
\end{itemize}

In Table \ref{tab:diversity}, for example, different epochs (training time) are used to include the effect of such possible temporal diversity in the proposed evolutionary environment. This selection allows us to include such a possible scenario in the analysis.

\section{Conclusion}
This research has shown that our proposition of building a colony of AI with multi-model and mixture-model families of AI agents can lead to an AI system that resembles the behavior of a biological colony. As demonstrated by a limited set of simulations, we can see that the mapping of the pretrained VGG16, VGG19, and ResNet50 models to a fast learner AI agent, a detailed learner AI agent, and an organized learner AI agent helps to build a colony of AI that can make collective decisions by sharing their locally learned knowledge. The novel representation of the AI models using triplets and hierarchical triplets, together with Genetic Algorithms, opens up a new research direction to understand the knowledge mechanism of the AI models. The use of fast learner AI agents, detailed learner AI agents, and organized learner AI can also help us enhance the diversity and the quality of a colony of AI. Hence, the proposed concept of colony of AI can help advance the emerging research in multi-agent AI systems. The next step of our research is to perform additional simulations with more complex, real-world datasets and compare with traditional multi-agent AI systems to support our conclusions. This research will also be extended to develop an explainable colony of AI system. 

\bibliographystyle{unsrtnat}
\bibliography{aaai25}  %%% Uncomment this line and comment out the ``thebibliography'' section below to use the external .bib file (using bibtex) .

\end{document}